\documentclass[conference]{IEEEtran}
\IEEEoverridecommandlockouts
\usepackage{cite}
\usepackage{amsmath,amssymb,amsfonts}

\usepackage{algorithmic}
\usepackage{graphicx}
\usepackage{textcomp}
\usepackage[table]{xcolor}

\usepackage{url}
\usepackage{hyperref}
\usepackage{afterpage}

\def\BibTeX{{\rm B\kern-.05em{\sc i\kern-.025em b}\kern-.08em
    T\kern-.1667em\lower.7ex\hbox{E}\kern-.125emX}}
 
\newcommand{\PaperTitle}[1]{Honey, I Shrunk The Actor:#1 A Case Study on Preserving Performance with Smaller Actors in Actor-Critic RL} 
   
\makeatletter
\newcommand{\linebreakand}{%
  \end{@IEEEauthorhalign}
  \hfill\mbox{}\par
  \mbox{}\hfill\begin{@IEEEauthorhalign}
}
\makeatother

\begin{document}

\title{ \PaperTitle{\\}
}

\author{\IEEEauthorblockN{Siddharth Mysore}
\IEEEauthorblockA{\textit{Department of Computer Science} \\
\textit{Boston University}\\
Boston, U.S.A. \\
sidmys@bu.edu}
\and
\IEEEauthorblockN{Bassel El Mabsout}
\IEEEauthorblockA{\textit{Department of Computer Science} \\
\textit{Boston University}\\
Boston, U.S.A. \\
bmabsout@bu.edu}
\and
\IEEEauthorblockN{Renato Mancuso}
\IEEEauthorblockA{\textit{Department of Computer Science} \\
\textit{Boston University}\\
Boston, U.S.A. \\
rmancuso@bu.edu}
\linebreakand
\IEEEauthorblockN{Kate Saenko}
\IEEEauthorblockA{\textit{Department of Computer Science, Boston University} \\
\textit{Co-affiliated with MIT-IBM Watson AI Lab}\\
Boston, U.S.A. \\
saenko@bu.edu}
}

\maketitle

\begin{abstract}
    Actors and critics in actor-critic reinforcement learning algorithms are functionally separate, yet they often use the same network architectures.
    This case study explores the performance impact of network sizes when considering actor and critic architectures independently. 
    By relaxing the assumption of architectural symmetry, it is often possible for smaller actors to achieve comparable policy performance to their symmetric counterparts. 
    Our experiments show up to 99\% reduction in the number of network weights with an average reduction of 77\% over multiple actor-critic algorithms on 9 independent tasks.
    Given that reducing actor complexity results in a direct reduction of run-time inference cost, we believe configurations of actors and critics are aspects of actor-critic design that deserve to be considered independently, particularly in resource-constrained applications or when deploying multiple actors simultaneously.
\end{abstract}


\section{Introduction}
    As a tool for developing machine learning (ML) based artificial intelligence (AI) for games, deep reinforcement learning (RL) has received increased interest in recent years.
    RL algorithms typically attempt to train control policies that maximize expected rewards on a given task and have been used to develop game-play AI ranging from human-like~\cite{autotesting, winningNotEverything,zhao2019multi} to superhuman~\cite{openai5,vinyals2019grandmaster}, with a wide array of applications, including game testing~\cite{winningNotEverything,autotesting,zhao2019multi}, competition against human players~\cite{openai5,vinyals2019grandmaster}, and simply as a means to develop more diverse styles of in-game play~\cite{winningNotEverything, alonsodeepRLAAANav20}.
    While RL can enable learning models for interesting behavior, the models can be significantly more computationally expensive than classical controllers or heuristic-based algorithms, particularly if multiple different models need to be deployed. 
    Broader ML applications can take advantage of powerful or dedicated compute resources but run-time inference costs are an important consideration for games when budgeting for AI and rendering, especially when games are expected to run on a diverse set of possible hardware configurations.
    High compute costs can ultimately hinder the accessibility of RL tools in development and deployment and the cost of a deep RL model is proportional to model complexity.
    In exploring how changing model complexity affects learning performance, we discovered that, for a class of RL algorithms, actor-critic RL, the size and run-time cost of learned actors could often be significantly reduced, while preserving performance parity with larger networks.
    
    Given a reward signal, RL algorithms learn to estimate the values of visited states and actions and use the learned values to inform policy optimization.
    Actor-critic methods, currently amongst the most common class of model-free deep RL algorithms, are characterized by their separation of the functions representing the actor (the RL policy) and the critic (the value function estimator).
    This allows actor-critic methods to theoretically take advantage of the improved sample efficiency of value-based RL approaches like Q-learning~\cite{Qlearning} while also capitalizing on the ability to use policy-based approaches such as policy gradient~\cite{sutton2000policy}, enabling learning on continuous action domains and improved robustness to stochasticity.
    
    The separation of actors and critics means that only the actor is required during inference, which saves the run-time cost of estimating policy value, as it is mainly useful only during training.
    Beyond excluding the critic, further reducing run-time compute requires reducing actor sizes, leading us to ask how small actors can get before losing learning efficacy.
    We noted that works considering the impact of network architectures on RL performance typically adjust both actor and critic sizes equivalently~\cite{pineau2020improving, Overfitting, Overfitting2, islam2017reproducibility, benchmarkingRL}, i.e. their architectures are kept `symmetric'. 
    We also see this implicit assumption in game-development tools such as Unity's ML-agents API~\cite{juliani2018unity}.
    Nonetheless, nothing in the theoretical foundations precludes using different architectures for actors and critics.
    In this work, we study if symmetry is important to actor-critic methods and how relaxing the self-imposed symmetry impacts performance and model complexity.
    We find this actor-critic architectural `asymmetry' allows for a significant reduction in actor network sizes without compromising the algorithms' performances.
    
    We hypothesize that modeling the value function associated with an environment often requires a higher capacity for modeling complexity. 
    This is because the critic needs to develop an understanding of both the dynamics of a black-box system and how they contribute to learning rewards.
    Actors just try to maximize the value, which can be done by gradient ascent on the value function estimate.
    Our results suggest this may often be easier to optimize, requiring less modeling complexity, allowing for smaller actors.
    We test 4 popular actor-critic algorithms on 9 different environments of varying system dynamics and complexity.
    We show that it is not simply that network architectures often presented in contemporary literature have a lot of headroom with model complexity but that actors often require less modeling capacity than critics, allowing them to function with smaller architectures.
    The practical implications of this intuitive observation have received little attention, both in the game AI and broader ML communities, and no prior study has been conducted to systematically understand and evaluate the potential of this architectural asymmetry in actor-critic methods.
    Our results show that architectural asymmetry enables reductions in actor sizes in excess of 98\% even over actors that have already been shrunk through symmetric architecture tuning, with an average reduction of 77\% in model parameters.

\section{Preliminaries}\label{section:bg}

    Actor-critic methods separate the actor and critic functions of an RL algorithm. 
    The actor function represents the policy, i.e. the core of the decision making aspect of the RL algorithm, while the critic estimates the value of trajectories of state transitions under the policy.
    Actor-critic algorithms are typically trained through sampling-based value iteration and policy gradients, where critics are trained to minimize the loss on the measured returns against the estimated value and actors are trained to maximize the estimated value of the trajectories generated under the actor policy.
    Prominent contemporary algorithms include A2C~\cite{A2C}, A3C~\cite{A3C}, PPO~\cite{PPO}, TRPO~\cite{TRPO}, DDPG~\cite{DDPG}, TD3~\cite{TD3}, and SAC~\cite{SAC}, to name a few.
    
    Network architecture is one of many hyper-parameters that can impact actor-critic performance~\cite{henderson2018deep, islam2017reproducibility, self-tuning, Overfitting, andrychowicz2021what, sinha2020d2rl}, but actors and critics are rarely considered separately.
    This can be an important to consider, but is easily missed given that most commonly available tools typically hard-code actor/critic symmetry. 
    Actor-critic algorithms are widely represented in common baselines and benchmarks, but when surveying existing baseline codes including OpenAI Baselines~\cite{baselines}, OpenAI Spinning Up~\cite{SpinningUp2018}, Stable Baselines (v2 and v3)~\cite{stable-baselines, stable-baselines3}, Unity ML-agents~\cite{juliani2018unity}, Tensorflow RL Agents~\cite{TFAgents}, and rllab~\cite{benchmarkingRL}, we found that actor and critic architectures are generally entangled and symmetric, often allowing users to define architectures for actor and critic networks together but not independently.
    There are many aspects to network architecture, including input handling, layer sizes, network depths, recurrence, information sharing between nodes, etc.
    While each aspect deserves attention, we focused on layer sizes while keeping all other parameters fixed as per values found in benchmarks for 2-hidden-layer networks.
    
\section{Litmus Test: A Toy Problem}\label{section:toy}
    
    \begin{figure*}[h]
        \centering
        \includegraphics[width=\textwidth]{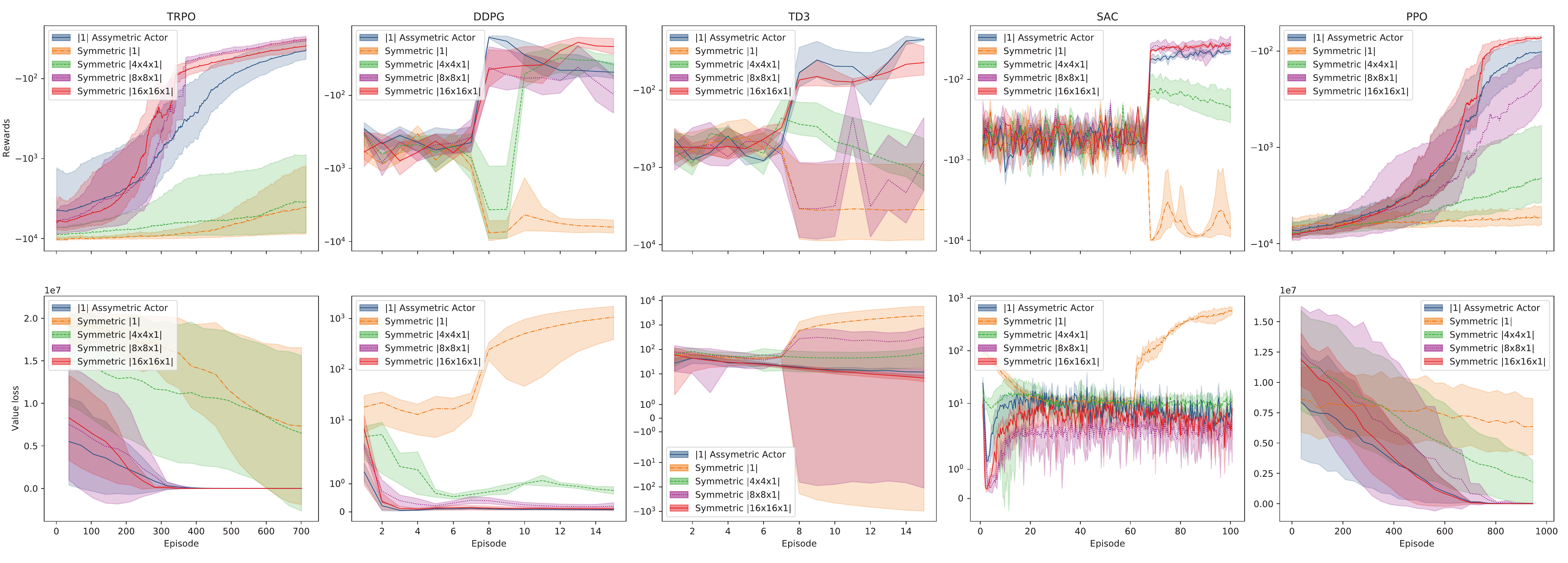}
        \caption{Comparison of how rewards and value (critic) losses evolve during training on a simple toy-problem. Results suggest a strong correlation between critic size and policy viability. Symmetric policies share the same size as their critics while the asymmetric actor has a 1-neuron policy and a $|16,16,1|$ critic. Note that the asymmetric actors performs very similarly to the symmetric $|16,16,1|$ policies and share similar value loss profiles too.}
        \label{fig:toy_losses}
    \end{figure*}
    
    \begin{figure*}[t]
        \centering
        \includegraphics[width=\textwidth]{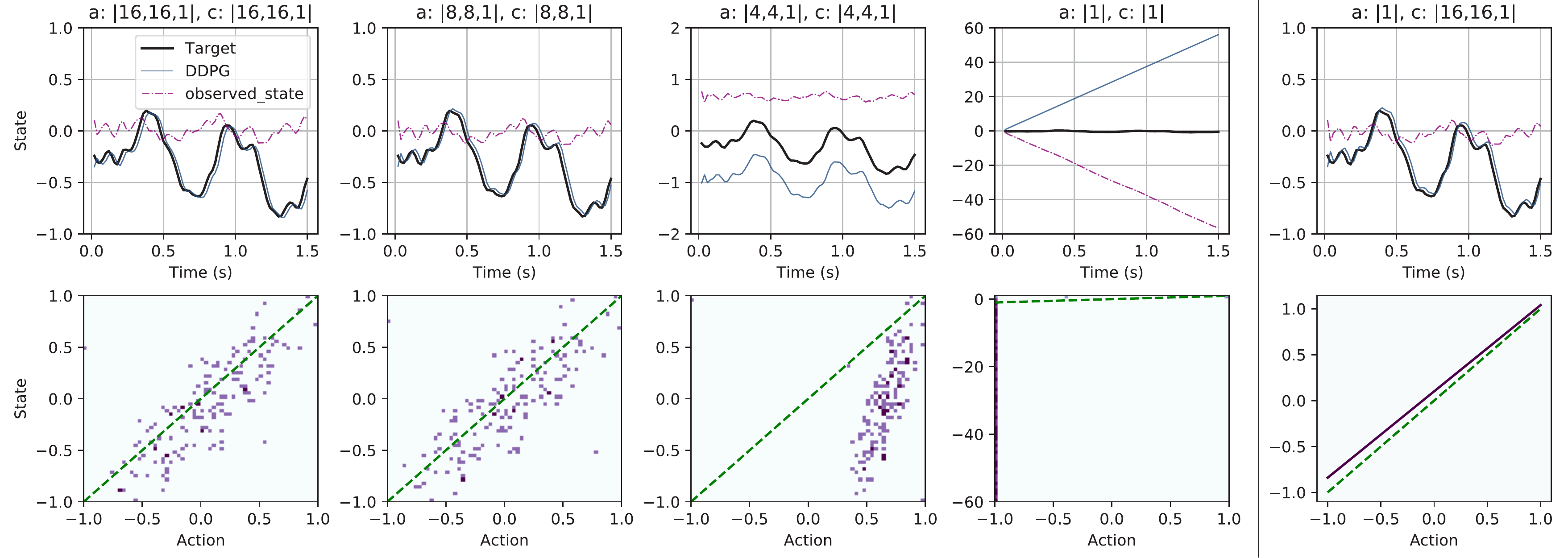}
        \vspace{-1.5\baselineskip}
        \caption{
        {Representative samples of DDPG behavior for different actor ``a" and critic ``c" sizes. 
        This visualizes the ability of differently configured agents, or lack thereof, to solve the simple toy problem described in Section~\ref{section:toy}.
        The upper row represents the state tracking problem and how well different sample agents performed in tracking a target trajectory.
        The lower row visualizes the state-action distributions for the same agents with green dotted line representing the ideal action policy for this problem, i.e. action $a$ = observed state $o$.
        Crucially, only the asymmetric linear actor (right) is able to learn a near-ideal policy, while its symmetric counterpart simply fails to learn and larger networks learn more (needlessly) complex policies.
        This shows that, with sufficient modeling capacity in the critic, architectural asymmetry allows a single actor neuron to solve this simple problem, where symmetry would impede it. }}
        \label{fig:toy_stateAction}
        \vspace{-1\baselineskip}
    \end{figure*}

    We start with a toy-problem that we know could be solved, in principle, by a very small, linear (1-neuron) actor.
    We constructed a simple one-dimensional goal-tracking problem without complex dynamics or system noise (details in Appendix~\ref{A:toy_setup}).
    The agent's observed state $o$ is the difference between the desired goal $g$ and current internal state $s$, $o = g - s$, where $s, g \in [-1, 1]$.
    Actions $a$ affect the internal state such that the next state $s' = s + a$, while the observed reward $r = -|g-s'|$.
    This problem is solved with the ideal policy $a^* = o$, which can be represented by a single neuron.
    
    We considered every algorithm included in the OpenAI~Spinning~Up code-base~\cite{SpinningUp2018}, which includes TRPO, PPO, DDPG, SAC and TD3.
    The code-base, by default, allows users to jointly adjust the number of hidden layers and the number of hidden neurons per layer in the actor and critic networks.
    For all algorithms tested however, when trained with the implicit assumption of actor and critic network symmetry, it was not possible for agents to learn to solve even this simple problem with a 1-neuron policy.
    We will note however that it is possible for the single neuron weight to be randomly initialized close to 1, which effectively solves the problem, though does not count as a `learned' solution.
    We tested network architectures including $|1|$, $|4,4,1|$, $|8,8,1|$ and $|16,16,1|$, where numbers within the vertical brackets provide a comma-delimited representation of the number of neurons per layer in the neural network, with the last layer being the 1-D output.
    When structured symmetrically, we found that agents did not consistently train to solve this problem for networks smaller than $|8,8,1|$, but were most stable with $|16,16,1|$.
    Fig.~\ref{fig:toy_losses} shows training trends for each of the algorithms trained on this problem while Fig.~\ref{fig:toy_stateAction} compares the quality of learned control.
    
    Note that a linear 1-neuron actor policy is not capable of learning to solve this problem when paired with a 1-neuron critic, but learns viable policies when trained with a larger critic.
    Trends in rewards and losses during training indicated that an improvement in value estimation was followed shortly thereafter by an improvement in the observed rewards.
    Based on this observation, we hypothesized that it was the smaller critics that were lacking in their ability to model the underlying value functions.
    To test this, we restricted the actor size to be 1-neuron and set the critic to be $|16,16,1|$.
    By breaking from the implicit assumption of actor-critic symmetry and separately considering the actor and critic architecture, it was possible to consistently train 1-neuron actors where the smaller symmetrically sized 1-neuron critics had failed.
    Figs.~\ref{fig:toy_losses}~and~\ref{fig:toy_stateAction} demonstrate that the asymmetric models share similar training and loss optimization characteristics to their larger symmetric counterparts while still allowing for successful training.

\section{Architectural-Asymmetry Benchmarked}

    Separately defining actor and critic architectures proved helpful in the case of a simple toy problem.
    To better understand the practical implications of actor-critic asymmetry, we evaluated how different network sizes affect performance on a number of common benchmark tasks from OpenAI's Gym~\cite{GYM} benchmark environments, in addition to games from the Pygame Learning Environment (PLE)~\cite{tasfi2016PLE} and Unity's ML-agents~\cite{juliani2018unity} example training environments\footnote{Training code and extended results available at \url{http://ai.bu.edu/littleActor/}}.
    
    To better characterize how network size can affect performance, and specifically how smaller networks impact performance, we consider two aspects of the actor-critic architecture: (i) the smallest symmetric actor-critic architecture that can meaningfully solve the problem, and similarly (ii) the smallest asymmetric architecture with comparable performance.
    Treating $|400,300|$ (the DDPG and TD3 default) as the upper bound on a set of possible 2-hidden-layer architectures, we consider the hidden-layer structures: $|1,1|$, $|4,4|$, $|8,8|$, $|16,16|$, $|32,32|$, $|64,64|$, $|128,128|$, $|256,256|$ and $|400,300|$ (note that these do not include the output layer, which is defined with respect to the action space for each test environment).
    This list covers typical architectures found in actor-critic literature but also goes smaller to get a sense the lower bound on the network complexity needed for the considered tasks.
    
    \begin{figure*}[t]
        \centering
        \includegraphics[width=\textwidth]{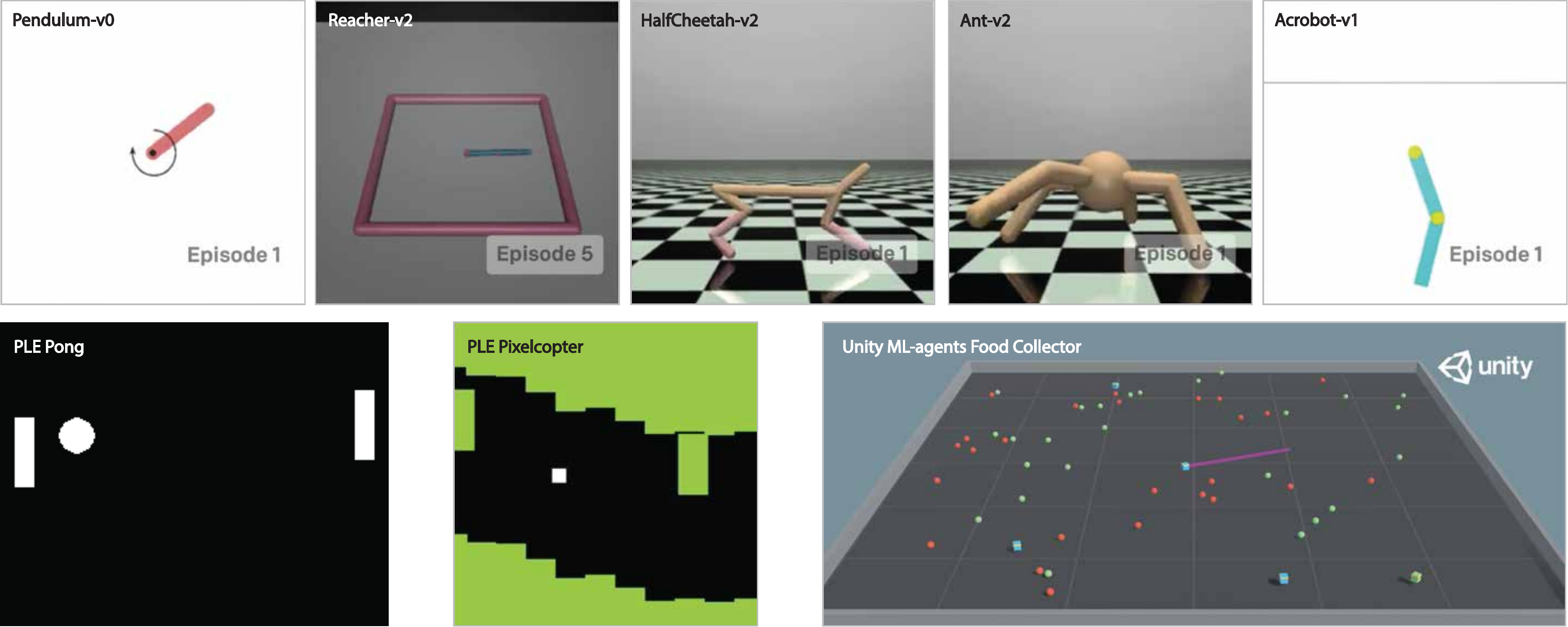}
        \caption{Environments used to benchmark various actor-critic algorithms. Environments were chosen to cover a diverse set of dynamics and objectives.}
        \label{fig:envs}
    \end{figure*}
    
    \afterpage{
    \begin{table*}[t!]
        \caption{Exploring the impact of actor-critic asymmetry on algorithm performance (improvements in size are in \bf{bold})}
        \label{table:Performance}
        \newcolumntype{g}{>{\columncolor{gray!30}}c}
        \newcolumntype{j}{>{\columncolor{gray!7}}c}
        \centering
        \setlength\tabcolsep{1.9pt}
        \begin{tabular}{c|||c||j|c||g|c||g|g|c|c}
            \hline
            {Algorithm} & Threshold & \multicolumn{2}{c||}{Baseline} & \multicolumn{2}{c||}{Symmetric Actor}  & \multicolumn{4}{c}{Asymmetric Actor}  \\ \cline{3-10}
            & & Size & Reward & Size $\downarrow$ & Reward $\uparrow$ & Actor Size $\downarrow$ & Reduction $\uparrow$ & Critic Size & Reward $\uparrow$ \\ \hline \hline
            \multicolumn{10}{c}{Gym Pendulum-v0} \\ \hline
            DDPG & -160 & $|400,300|$ & $-145.56 \pm 10.64$  & $ |16,16| $ & $-150.28 \pm 9.08$ &   $\bf|4,4|$ & $\bf88.38\%$ & $|16,16|$ & $-158.97 \pm 5.33$ \\
            TD3  & -160 & $|400,300|$ & $-152.71 \pm 9.47$   & $ |16,16| $ & $-152.42 \pm 6.65$ &   $\bf|4,4|$ & $\bf88.38\%$ & $|16,16|$ & $-167.57 \pm 14.53$ \\
            SAC  & -160 & $|256,256|$ & $-139.86 \pm 8.29$   & $ |16,16| $ & $-155.32 \pm 11.25$ &  $\bf|4,4|$ & $\bf88.38\%$ & $|16,16|$ &$-153.86 \pm 10.97$ \\
            PPO  & -200 & $|64, 64| $ & $-668.60 \pm 551.85$ & $|128,128|$ & $-193.55 \pm 40.14$ & $|128,128|$ & $0\%$ & $|128,128|$ & $-193.55 \pm 40.14$ \\ \hline
            \multicolumn{10}{c}{Gym Reacher-v2} \\ \hline
            DDPG & -5.0 & $|400,300|$ & $-4.26 \pm 0.25$ & $ |64,64|$   & $-4.75 \pm 0.19$ & $\bf|8,8|$ & $\bf96.33\%$ &  $|64,64|$ & $-4.80 \pm 0.44$ \\
            TD3  & -7.0 & $|400,300|$ & $-6.52 \pm 1.12$ & $|64,64| $ & $-6.91 \pm 0.74$ & $\bf|32,32|$ & $\bf70.23\%$ & $|64,64|$ & $-6.68 \pm 1.21$ \\
            SAC  & -6.5 & $|256,256|$ & $-5.96 \pm 0.47$ & $|128,128|$  & $-6.05 \pm 0.91$ & $\bf|16,16|$ & $\bf97.28\%$ & $|128,128|$ & $-6.02 \pm 1.07$ \\
            PPO  & -5.5 & $|64, 64| $ & $-4.37 \pm 1.74$ & $ |64,64|$  & $-4.37 \pm 1.74$ & $\bf|16,16|$ & $\bf90.15\%$ & $|64,64|$ &$-5.49 \pm 1.00$ \\ \hline
            \multicolumn{10}{c}{Gym HalfCheetah-v2} \\ \hline
            DDPG & 7000  & $|400,300|$ & $7026.01 \pm 202.78$ & $  |64,64| $ & $7450.01 \pm 950.15$ & $\bf|32,32|$ & $\bf68.01\%$ & $|64,64|$ & $8273.76 \pm 437.66$ \\
            TD3  & 8000  & $|400,300|$ & $8861.92 \pm 870.02$ & $ |64,64| $ & $8315.13 \pm 262.78$ & $\bf|32,32|$ & $\bf68.01\%$ & $|64,64|$ & $8145.84 \pm 262.55$ \\
            SAC  & 10000 & $|256,256|$ & $11554.76 \pm 779.91$ & $ |64,64|$ & $10180 \pm 759.10$ & $\bf|32,32|$ & $\bf68.01\%$ & $|64,64|$ & $9619 \pm 158.40$ \\
            PPO  & 3000  & $|64, 64| $ & $3395 \pm 1156.30$ & $|64, 64|  $ & $3395 \pm 1156.30$ & $\bf|32,32|$ & $\bf68.01\%$ & $|64,64|$ & $3089 \pm 919.25$ \\ \hline
            \multicolumn{10}{c}{Gym Ant-v2} \\ \hline
            DDPG$^\mathrm{a}$ &  --  & $|400,300|$ & $225.23 \pm 362.88$  & --  & -- & -- & -- & -- \\
            TD3  & 3000 & $|400,300|$ & $3087.86 \pm 888.75$ & $|256,256|$ & $3944.78 \pm 745.48$ & $\bf|32,32|$ & $\bf94.92\%$ & $|256,256|$ & $3553.52 \pm 396.30$ \\
            SAC  & 3000 & $|256,256|$ & $3366.07 \pm 1522.45$ & $ |64,64|$ & $3108.63 \pm 519.59$ & $|64,64|$ & $0\%$ & $|64,64|$ & $3108.63 \pm 519.59$ \\
            PPO  & 3000 & $|64, 64| $ & $3734.58 \pm 988.29$ &  $ |8,8|  $ & $3723.57 \pm 760.94$ & $|8,8|$   & $0\%$ & $ |8,8|  $& $3723.57 \pm 760.94$ \\ \hline
            \multicolumn{10}{c}{Gym Acrobot-v1} \\ \hline
            SAC$^\mathrm{b}$ & -100 & $|256,256|$ & $-76.5 \pm 4.30$ & $|32,32|$ & $-86.5 \pm 7.54$ & $\bf|16,16|$ & $\bf68.46\%$ & $|32,32|$ & $-90.5 \pm 8.12$ \\
            PPO$^\mathrm{b}$ & -100 & $|64,64|$ & $-74.7 \pm 0.30$ & $|8,8|$ & $-70.75 \pm 0.33$ & $\bf|1,1|$ & $\bf90.33\%$ & $|8,8|$ & $-71.75 \pm 0.50$ \\ \hline
            \multicolumn{10}{c}{PLE Pong} \\ \hline
            SAC$^\mathrm{b}$ & 0.90 & $|256,256|$ & $0.93 \pm 0.05$ & $|64,64|$ & $0.90 \pm 0.06$ & $\bf|16,16|$ & $\bf90.73\%$ & $|64,64|$ & $0.89 \pm 0.03$ \\
            PPO$^\mathrm{b}$ & 0.80 & $|64,64|$ & $0.85 \pm 0.13$ & $|64,64|$ & $0.85 \pm 0.13$ & $\bf|4,4|$ & $\bf98.62\%$ & $|64,64|$ & $0.83 \pm 0.09$ \\ \hline
            \multicolumn{10}{c}{PLE Pixelcopter} \\ \hline
            SAC$^\mathrm{b}$ & 35 & $|256,256|$ & $36.83 \pm 9.65$ & $|64,64|$ & $33.87 \pm 5.80$ & $\bf|8,8|$ & $\bf96.79\%$ & $|64,64|$ & $35.98 \pm 4.93$ \\
            PPO$^\mathrm{b}$ & 20 & $|64,64|$ & $22.24 \pm 9.86$ & $|64,64|$ & $22.24 \pm 9.86$ & $\bf|32,32|$ & $\bf71.30\%$ & $|64,64|$ & $19.91 \pm 10.03$ \\ \hline
            \multicolumn{10}{c}{Unity ML-agents Food Collector} \\ \hline
            SAC$^\mathrm{c,d}$ & 60 & $|128,128|^\mathrm{e}$ & $61.77 \pm 6.16$ & $|32,32|$ & $65.22 \pm 8.26$ & $\bf|8,8|$ & $\bf81.48\%$ & $|32,32|$ & $60.31 \pm 5.49$ \\
            PPO$^\mathrm{c}$ & 50 & $|128,128|^\mathrm{e}$ & $50.25 \pm 2.01$ & $|128,128|$ & $50.25 \pm 2.01$ & $\bf|32,32|$ & $\bf87.82\%$ & $|128,128|$ & $48.05 \pm 3.12$ \\ \hline
            \multicolumn{10}{l}{$^\mathrm{a}$Due to DDPG being incapable of reasonably solving the Ant task in our experiments, it is omitted for the Ant-v2 environment.}\\
            \multicolumn{10}{l}{$^\mathrm{b}$OpenAI Spinning up implementation of algorithm is adjusted to support discrete action spaces.}\\
            \multicolumn{10}{l}{$^\mathrm{c}$Unity ML-agents implementation of algorithm is used to preserve compatibility with recommended baseline configuration.}\\
            \multicolumn{10}{l}{$^\mathrm{d}$Due to the instability of ML-agent's implementation of SAC's training on Food Collector, we report the average maximum reward achieved during training.}\\
            \multicolumn{10}{l}{$^\mathrm{e}$We use as a baseline the sizes recommended by the ML-agents baseline configuration parameters.}
        \end{tabular}
    \end{table*}
    \begin{figure*}[h!]
        \centering
        \includegraphics[width=\textwidth]{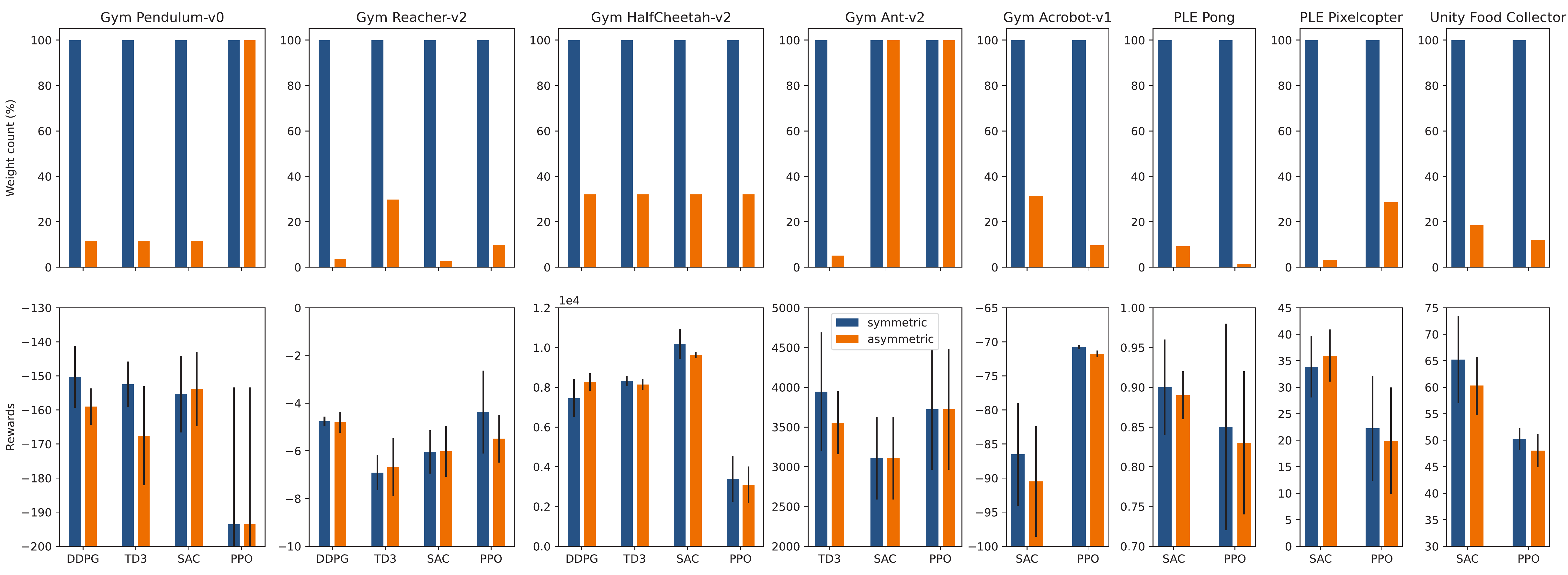}
        \caption{
            Visual representation of the relative network size reductions and rewards presented in Table~\ref{table:Performance}. Network sizes are computed per task (see Appendix~\ref{A:net_sizes}). Results evidence that it is often possible to train significantly smaller actors with larger critics without compromising performance by breaking implicit assumptions of network architectural symmetry between actors and critics. 
        }
        \label{fig:relative_perf}
    \end{figure*}
    }
    
    \emph{Experimental Method}:
    We determine appropriate sizes for the smallest symmetric and asymmetric architectures sequentially.
    By running the algorithms on 6 random seeds with their originally reported architectures, we establish a baseline and a threshold target performance for smaller networks.
    We then use this threshold, with a 10\% tolerance on the lower-bound, to perform a binary search over the actor-critic network sizes to find the smallest one that can solve the problem while preserving actor-critic symmetry, similarly averaging performance over 6 seeds.
    Finally, by locking the critic architecture to the smallest symmetric size found, we repeat the binary search over the possible actor architectures to find the smallest asymmetric actor that would solve the problem.
    This allows us to determine more conclusively that, in cases where asymmetry allows for actor size reduction, it is not simply attributable to excess modeling capacity in both the actor and critic networks when using baseline network architectures.
    
    \begin{figure*}[t]
        \centering
        \includegraphics[width=\textwidth]{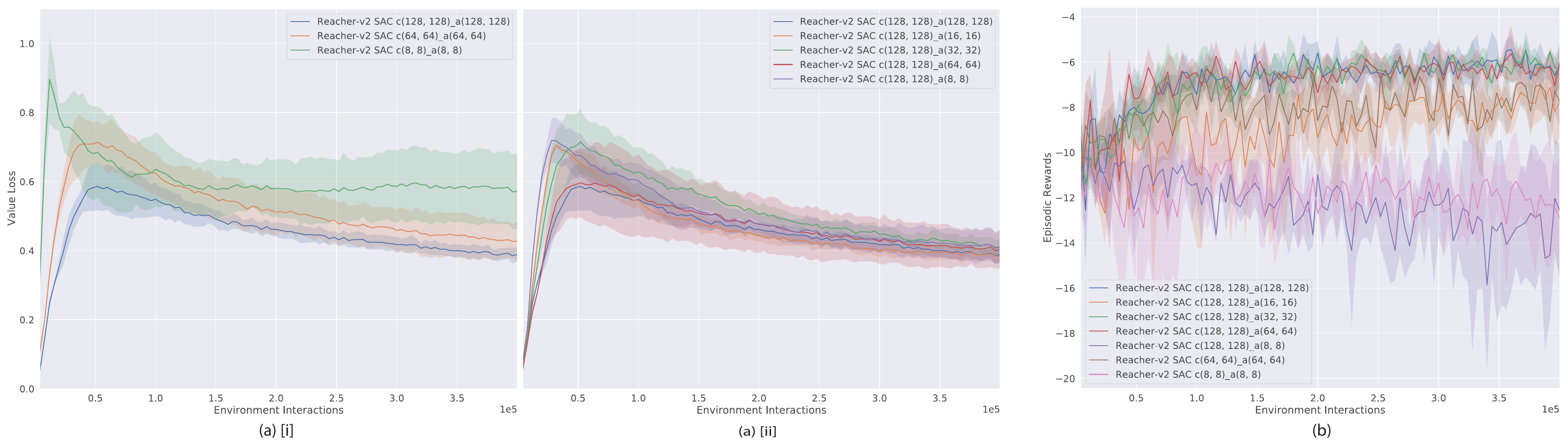}
        \vspace{-1.5\baselineskip}
        \caption{Training progress for SAC on Reacher-v2: (a) Q-Value Loss during training and (b) the episodic returns.
        (a)[i] Shows the value loss during training for different critic sizes while (a)[ii] shows the same for different actors trained with the same critic size.
        Changing the critic's size impacts value-loss more than changing the actor's, suggesting that the critic is more sensitive to changes in modeling capacity.}
        \label{fig:reacher_sac}
        \vspace{-1\baselineskip}
    \end{figure*}
    
    \emph{Learning Tasks}:
    In an effort to cover a wide range of possible system dynamics, we experimented with 8 different training environments (Fig.~\ref{fig:envs}).
    From the Gym benchmarks, we tested controls tasks of varying levels of complexity: Pendulum-v0, Reacher-v2, Ant-v2, HalfCheetah-v2 and Acrobot-v1. 
    Agents have control over various actuators and use information observed about joint dynamics to control joint actuation.
    From PLE, we tested an implementation of Pong, where the RL agent competes against a heuristic-based opponent AI, as well as Pixelcopter, an obstacle-avoidance side-scroller.
    For both the PLE games, the agents directly observe relevant game-state information such as player position, obstacle/ball positions, player speed, etc. instead of relying on more complex information structures such as the video frame buffer or RAM state as found in some game benchmarks --- this was a deliberate choice to better reflect how an RL agent might more likely be practically implemented to play these games.
    Finally, we experimented with the Food Collector game from Unity ML-agents (release~14) set of example training environments, where the agent observes information about the game-state in a local region around its position and is tasked with collecting objective mobile markers using a designated capture action, while avoiding mobile obstacles.
    This game was chosen specifically as its construction bears relevance to a wide variety of objective-capture type games.
    While other games from the ML-agents examples were considered, we opted against using any that employed self-play in training to avoid behavioral artefacts that could arise and felt that the other environments did not offer significantly more representative value, given the controls tasks and games already tested.
    
    \emph{Algorithms Tested}:
    For the continuous control tasks, we tested 4 algorithms: DDPG~\cite{DDPG}, TD3~\cite{TD3}, SAC~\cite{SAC} and PPO~\cite{PPO}.
    With the exception of PPO, we used codes provided by the OpenAI Spinning Up~\cite{SpinningUp2018} code-base, with training hyper-parameters from the RL-zoo~\cite{stable-baselines}.
    As shown by Engstrom et al.~\cite{Engstrom2020Implementation} and Ilyas et al.~\cite{Ilyas2020A} PPO can be highly sensitive to implementation details, so we instead used the OpenAI Baselines~\cite{baselines} implementation of PPO to maintain parity with the original work.
    For the discrete action-space environments (Acrobot, Pong and Pixelcopter), DDPG and TD3 were not tested as they are designed specifically for continuous control.
    For Unity's Food Collector, we utilized the ML-agents implementations of SAC and PPO as adapting the Spinning Up codes to work with the API proved non-trivial.
    
    \emph{Results}:
    Our results are shown in Table~\ref{table:Performance}, with a visual contextualization of the relative performances of symmetrically and asymmetrically constructed actors shown in Fig.~\ref{fig:relative_perf}.
    Asymmetry has demonstrable utility in reducing actor sizes in every environment tested, though not necessarily with every algorithm in each environment.
    We would also note that, though it is possible for the automated search to find actors larger than their critics, such a case did not arise in testing.
    Network size reductions vary by algorithm and environment, with between 70--97\% actor size reductions, and with the Q-learning based algorithms (DDPG, TD3 and SAC) sharing similar trends.
    Network sizes are computed per task to account for the different state and action dimensions (see Appendix~\ref{A:net_sizes}). 
    
    PPO appears to be more unpredictable in performance, particularly when considering the disparities in network size to performance over Pendulum-v0, Ant-v2 and Acrobot-v1.
    Despite poor performance on the simple inverted pendulum problem with smaller networks where other algorithms succeed, PPO solves the more complicated multi-pedal locomotion problem in Ant-v2 with significantly smaller actor and critic networks compared to the other algorithms tested and is similarly able to solve the Acrobot-v1 problem with smaller networks than SAC.
    It does however allow actor size reduction through asymmetry in most of the tested environments.
    
    Experimental observations on these more complex environments also provide further empirical validation of our hypothesis that modeling capacity in the critic is often the limiting factor in achieving successful training.
    This is best illustrated by data on the training of SAC on the Reacher-v2 task and is shown in Fig.~\ref{fig:reacher_sac} where, much like in Fig.~\ref{fig:toy_losses}, there are relatively clear demarcations in where the Q-value losses and episodic rewards saturate for different network sizes.
    While changing the size of the critic during the symmetric binary search has a noticeable impact on the Q-value loss, changing the actor size during asymmetric search does not impact the Q-value loss significantly, as shown when comparing Figures~\ref{fig:reacher_sac}(a)[i] and [ii].
    A larger critic also allows for a much smaller actor to be successfully trained (in this case a critic with hidden layers of size $|128,128|$ allows for an actor of size $|16,16|$).
    We can therefore draw the conclusion that the dominant limiting factor in learning, for this algorithm, on this environment, is indeed the modeling capacity of the critic.
    Similar trends were observed for other algorithms in other training tasks too.
    
    Our project website:~\url{http://ai.bu.edu/littleActor/} provides access to view full training data for the experiments summarized in Table~\ref{table:Performance} and Figs.~\ref{fig:relative_perf}~and~\ref{fig:reacher_sac}, in addition to training code used.

\section{Reflection and Recommendations}
    The main goal of this work was to draw attention to the practical benefits of separately considering actor and critic architectures in RL-based AI, and experiments over a wide array of tasks show that large reductions in actor-network sizes are often possible.
    Furthermore, these reductions come without incurring significant degradation in performance (capped here to 10\%) --- and crucially, the tasks are still solved.
    We do recognize that, on the scale of typical `large' neural network models, the architectures considered in this paper could all be considered relatively small.
    However, with applications such as games, where running the game AI is just one of many compute tasks that need to be managed on machines with varying resources budgets and where it may be necessary to deploy multiple independent AIs simultaneously, we believe that network size can be an important optimization consideration.
    Reducing actor network sizes was also observed to have some benefit in reducing training time --- presenting with up to 50\% reduction in training time --- though these were inconsistent due to the impact of other tasks running concurrently on the training machines.
    
    While we made efforts to thoroughly explore one aspect of network architectures, i.e. the network size, as the network size is typically the largest contributor to compute cost, this should not be regarded as an exhaustive study of the impact of network architectures. 
    There are still more aspects such as depth, output activation, network initialization, etc., and as shown by Henderson et al.~\cite{henderson2018deep} and Andrychowicz et al.~\cite{andrychowicz2021what}, those can strongly contribute to policy performance.
    The key take-aways are that these aspects of network design bear serious consideration when developing AI for deployment, and while our analysis has been focused on network size, we suspect that it would be worth considering other architectural design choices separately for actors and critics as well.
    We would recommend that practitioners pay specific attention to how policy rewards and value losses evolve as a function of these architectural choices.
    If starting from symmetric architectures, we would recommend identifying the point of diminishing returns when increasing the critic parameters and building from there to reduce actor complexity.

\section{Related Work}

    Prior work by Islam et al.~\cite{islam2017reproducibility} and Henderson et al.~\cite{henderson2018deep} touch on the impact of network size on the performance of RL algorithms but their analyses are more limited to a smaller set of environments and architectures.
    Both works consider networks with 3 different possible hidden layer configurations: $|400, 300|$, $|64,64|$ and $|100, 50, 25|$, common sizes used in prior literature.
    There is anecdotal evidence in these works to suggest that larger actors are not necessarily more performant and that larger critics can usually learn better, but the data presented does not lend itself to clear conclusions.
    
    Asymmetry in actors and critics appears to have first been explicitly addressed by Pinto et al.~\cite{pinto2017asymmetric} in controlling robotic manipulation.
    The authors explore a form of input asymmetry between actors and critics, where actor networks are offered reduced state information from an RGBD sensor, while the critic is offered full-state information from a simulator during training.
    They demonstrate that actors are able to actually perform better under this reduced information paradigm, likely due to the reduced representational complexity.
    An interpretation of the implications of their work is that it also hints at the idea of the onerous of learning more robust and complex representations may perhaps lie more with the critic side of the network.
    
    Henderson et al.~\cite{henderson2018deep} provide a cursory analysis on the impact of disentangling the actor and critic architectures and consider how different algorithms behave in different training environments as a result of changing network structures.
    They note the sensitivity of algorithms to these details but a limitation of their analysis is that it sacrifices depth for breadth as this is just one of many hyper-parameters they investigate. 
    Their analysis appears to primarily seek to identify \emph{if} there is a correlation between performance and actor and critic architectures when considered independently.
    However only two environments and a limited set of network configurations are tested, preventing clear conclusions being drawn on \emph{how} actor and critic architectures relate to each other.
    By considering a wider range of network sizes and environments, we are able to formulate and test a clearer hypothesis on the relationship between network sizes, both for the actors and critics, and their impact on performance.
    
    Recent concurrent work by Andrychowicz et al.~\cite{andrychowicz2021what} discusses a large scale study on the impact of different hyperparameters on RL performance on 5 OpenAI Gym~\cite{GYM} continuous controls benchmark tasks.
    Network architecture is among the many parameters considered and, like us, the authors also extend their study past the typically used architectures to consider a wider range of network sizes, the impact of varying actor and critic sizes separately and also of varying the depth of the networks.
    Like us, the authors make note of the fact that it is possible to use smaller actors and that RL agents more often benefit from having larger critics.
    They also identified that increasing network depth was not necessarily beneficial, an observation also noted by Sinha et al.~\cite{sinha2020d2rl}.
    The authors' focus, being on performance maximization, as opposed to network complexity reduction however, results in them ultimately prioritizing different aspects of the network architecture --- such as policy initialization and output activation --- in addition to a wider array of non-architecture-related parameters.
    While they do contribute useful and important observations and recommendations on tuning those parameters for more performant policies, they do not make stronger statements on how the sizes of actors and critics may interact with each other, but mainly that they ought to be paid attention to.
    Their results do however independently corroborate ours (and ours, theirs).
    
    Reducing the parameter count of neural networks has also received interest in the broader machine learning community.
    Techniques like pruning~\cite{Han2016DeepCC},  knowledge distillation~\cite{Hinton2015DistillingTK}, and weight sharing~\cite{Ullrich2017SoftWF} have been shown to accelerate inference and reduce memory requirements, enabling deployment on resource-constrained mobile hardware.
    In RL, policy distillation~\cite{rusu2015policy} has been demonstrated as an effective tool for reducing network sizes using a teacher-student method where a larger pre-trained teacher network guides a smaller student network which learns to match the teacher's performance.
    Smaller students learning equivalently functional policies suggests an excess in modeling capacity of more typical architectures for policy networks and inspired us to question if it would be possibly to train smaller networks directly, without first needing to train a larger teacher.

\section{Conclusion}

    We hypothesized that this capacity for actor size reduction comes from the burden of modeling and understanding environment dynamics falling largely to the critic, rather than the actor.
    Provided that the critic has sufficient capacity to learn a decent estimation of the value function, we predicted that it would often be possible to train actors that are significantly smaller than the critic.
    The results presented in this case study substantiate this hypothesis.
    We demonstrated that, by relaxing the implicit assumption of symmetry in the architectures of actors and critics in actor-critic RL, it is possible to train significantly smaller networks for tasks without a significant degradation in policy performance.
    We demonstrate as much as a 97\% reduction in the number of weights needed to represent viable actor policies, with an average actor size reduction of 77\% 
    in actor network sizes for the tested tasks.
    We believe that the implications of these results can be highly significant to practical applications of RL, particularly in resource-constrained systems, and that it is also worth being generally aware of as part of the broader effort to understand how network sizes can impact actor-critic RL performance.\\\\
    \noindent \textbf{Acknowledgements:} This work was supported in part by grants NSF 1724237 and NSF CCF 2008799.
    
\bibliographystyle{IEEEtran}
\bibliography{references.bib}

\appendix
\counterwithin{table}{section}
\subsection{Toy Problem Setup Details}\label{A:toy_setup}
    \noindent Additional setup details for the toy-problem in Section~\ref{section:toy}:
    \begin{itemize}
        \item   {\bf Goal generation}: Goals $g_t$ are procedurally generated by functions mapping an input `time' signal to the range $[-1,1]$ using a simplex~\cite{noise} function.
        \item   {\bf System Dynamics}: The response to action,~$a_t$ is computed as $s_{t+1} = \text{clip}(s_t + a_t, -1, 1)$.
    \end{itemize}
        
\subsection{Per-task Policy Network Sizes}\label{A:net_sizes}
    \noindent Weight-counts for each environment are computed separately to account for their different state and action dimensions.
    \begin{table}[h]
        \caption{Number of actor-network weight parameters per environment for each of the environments considered}
        \label{table:net_sizes}
        \centering
        \setlength\tabcolsep{2pt}
        \begin{scriptsize}
        \begin{tabular}{c||c|c|c|c|c|c|c|c}
            \hline
            {Hidden Size} & Pen & Rea & HCh & Ant & Acr & Png & Pxl & FdC \\\hline
            $|400,300|$ & 122201 	& 125702 	& 129306	& 167508	& 124003 & 124403	& 124102	& 143104	\\
            $|256,256|$ & 67073 	& 69378 	& 71942		& 96520		& 68355	 & 68611	& 68354		& 80644		\\
            $|128,128|$ & 17153 	& 18306 	& 19590		& 31880		& 17795	 & 17923	& 17794		& 23940		\\
            $|64,64|$ 	& 4481 		& 5058 		& 5702		& 11848		& 4803	 & 4867		& 4802		& 7876		\\
            $|32,32|$ 	& 1217 		& 1506 		& 1830		& 4904		& 1379	 & 1411		& 1378		& 2916		\\
            $|16,16|$ 	& 353 		& 498 		& 662		& 2200		& 435	 & 451		& 434		& 1204		\\
            $|8,8|$ 	& 113 		& 186 		& 270		& 1040		& 155	 & 163		& 154		& 540		\\
            $|4,4|$ 	& 41 		& 78 		& 122		& 508		& 63	 & 67		& 62		& 256		\\
            $|1,1|$ 	& 8 		& 18 		& 32		& 130		& 15	 & - 		& - 		& -	        \\ \hline
            \multicolumn{1}{r}{Pen:} & \multicolumn{8}{l}{Gym Pendulum-v0}\\
            \multicolumn{1}{r}{Rea:} & \multicolumn{8}{l}{Gym Reacher-v2}\\
            \multicolumn{1}{r}{HCh:} & \multicolumn{8}{l}{Gym HalfCheetah-v2}\\
            \multicolumn{1}{r}{Ant:} & \multicolumn{8}{l}{Gym Ant-v2}\\
            \multicolumn{1}{r}{Png:} & \multicolumn{8}{l}{PLE Pong}\\
            \multicolumn{1}{r}{Pxl:} & \multicolumn{8}{l}{PLE Pixelcopter}\\
            \multicolumn{1}{r}{FdC:} & \multicolumn{8}{l}{Unity ML-agents Food Collector}
        \end{tabular}
        \end{scriptsize}
    \end{table}


\end{document}